# A Survey on Neural Architecture Search Based on Reinforcement Learning

Wenzhu Shao*
Tianjin University of Science and Technology
Tianjin, China
xiangyoua@mail.tust.edu.cn

*Abstract*—The automation of feature extraction of machine learning has been successfully realized by the explosive development of deep learning. However, the structures and hyperparameters of deep neural network architectures also make huge difference on the performance in different tasks. The process of exploring optimal structures and hyperparameters often involves a lot of tedious human intervene. As a result, a legitimate question is to ask for the automation of searching for optimal network structures and hyperparameters. The work of automation of exploring optimal hyperparameters is done by Hyperparameter Optimization. Neural Architecture Search is aimed to automatically find the best network structure given specific tasks. In this paper, we firstly introduced the overall development of Neural Architecture Search and then focus mainly on providing an overall and understandable survey about Neural Architecture Search works that are relevant with reinforcement learning, including improvements and variants based on the hope of satisfying more complex structures and resource-insufficient environment.

## I. Introduction

In the last decade, the process of feature extraction in machine learning, which used to need a lot of expert work, has been automated by the development of deep learning. However, the design of network architecture still rely heavily on human's expertise and rich experience gained from numerous experiments. In deep learning, the hyperparameters play a significant role in the performance of the network. Many networks that are proposed in some novel papers are hard for other researchers to implement. One of the most important reasons for that is it also requires much tedious fine-tuning work done by experts to explore the optimal hyperparameters for their networks. Hence, it is required that machine take on the task of designing better network architectures.

Before the deep learning became famous, the automation of searching the optimal hyperparameters mainly focused on the parameters in traditional machine learning algorithms. There existed some classical search strategies,such as random search, grid search, Bayesian optimization, reinforcement learning, and evolutionary algorithms. All of the strategies mentioned above are called as Hyperparameter optimization.

The hyperparameters in deep learning include training parameters and architecture parameters. The training parameters, such as learning rate, batch size and weight decay, define how to train a deep neural network. The architecture parameters define what the neural network consists of. For example, the numbers of layers, the type of each layer and filter size in convolutional networks, these parameters decide the structure of the network. The challenge is that these architecture hyperparameters are high-dimensional, discrete and mutual-related. The work of Neural Architecture Search (NAS) is trying to find the optimal architecture hyperparameters, even exploring novel architectures which have not been proposed by human researchers.

Since the researchers first proposed the concept of Neural Architecture Search, it has been a promising research project. The overall steps of Neural Architecture Search are shown in Fig. 1, and the descriptions are as follows:

*1) define a search space, in which, different network layers and operations are represented as some embedding such as strings and vectors.*

*2) apply a search strategy to search for the candidate network architectures.*

*3) build networks according to the search results and evaluate their performance in some tasks.*

*4) begin next search iteration based on the evaluation of the last generated networks.*

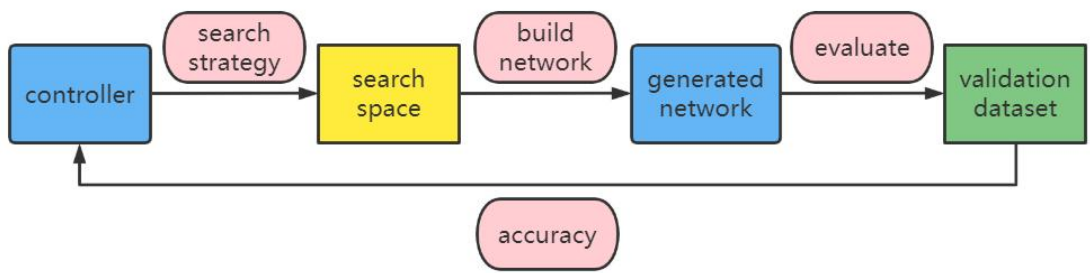

Fig. 1. The overall steps of neural architecture search

The definition of the search space is related with the development of deep learning research. In the beginning, the common convolutional neural networks are structured linearly, so the NAS only needed to consider how many layers there are, which type of operations of each layer and corresponding hyperparameters of each type of operations. The emerge of more complex deep neural networks such as ResNet, DenseNet and Skip connection led to the NAS to take considerations of multi-connection structures. Recently, the deep neural networks started to include repeated sub-structures called cells or blocks. The researchers of NAS proposed search based on cells, which means that NAS only search the structures of the cells and how to connect these cells [1][2][3].

Researchers have applied many optimization algorithms as search strategy in order to search for optimal architectures from the search space. For example, random search, Bayesian optimization, gradient-based method, reinforcement learning, evolutional algorithms and so on. Random search is the easiest method to understand and implement, so it has often been used as baseline. Despite of the simpleness of random search method, its performance is hard to be surpassed, especially when it is combined with some tricks like early-stopping [4]. Bayesian optimization achieved early successes in NAS since 2013 [5]. NAS became a mainstream popular research topic after researchers applied reinforcement learning as the search strategy to achieve competitive performance on the CIFAR-10 and Penn Treebank benchmarks in 2017 [6]. Evolutional algorithms were introduced by Google to solve NAS problems. It was also proved by Google that reinforcement learning and evolutional algorithms achieved the same accuracy which was slightly better than that of random search. But evolutional algorithms searched faster and generated smaller models than reinforcement learning did [7]. NAS applying reinforcement learning and evolutional algorithms search in discrete space and there does not exist a specific objective function. The objective function is regarded as a black box. If the search space is continuous instead of discrete, then the objective function is differentiable. It can provide convenience for gradient-based method to search the optimal parameters more effectively. Differentiable Architecture Search was thus proposed by researchers from CMU and Google [8]. The performance of gradient-based method is less satisfactory than that of reinforcement learning in NAS problems, while the latter normally costs huge computational resources.

The main challenge of Neural Architecture Search is that the implementation is costly. The search space of NAS is normally enormous, and it often costs huge computational resources to train the generated networks for evaluation. It means that the research of NAS was not suitable for common application. In order to solve this problem, many researches managed to accelerate the training of NAS, such as hierarchical representation, weight sharing and performance prediction.

Most experiments applied NAS to solve image classification tasks and natural language processing tasks. The main focus of these experiments is the accuracy of the generated models. In recent years, NAS has been applied in other tasks including semantic segmentation. Researchers started to take more factors into consideration about how to evaluate the performance of the generated architectures more comprehensively, which is called Multi-objective NAS.

This paper mainly focuses on providing general review on the research outputs about Neural Architecture Search which apply reinforcement learning as search strategy. In Section Ⅱ, researches about two classical NAS applying different reinforcement learning (RL) algorithms will be discussed. In Section Ⅲ, researches about improvement in the speed and costs of computation of NAS with RL will be discussed. In Section Ⅳ, more variants and extension based on NAS with RL will be discussed. Future work in this domain will be discussed in Section Ⅴ.

## Ⅱ. Classical Researches About Neural Architecture Based On Reinforcement Learning

In this section, different RL algorithms which have been incorporated in NAS as optimization methods will be discussed. Q-learning [9] is one of the most common algorithms that is used to solve neural architecture search, which is often combined with epsilon-greedy and experience replay. Q-learning was first used by researchers to propose MetaQNN which will be talked about with more detail in this section [10]. Q-learning was also incorporated to search in a cell-based search space [2], which will also be mentioned in section Ⅲ. Policy gradient methods also function as an alternative as optimization methods. The Neural Architecture Search going to be talked about in detail later in this section is the neural architecture search with reinforcement learning. It was one of the first two pioneering researches that tried to use reinforcement learning to solve neural architecture search problems. It used a policy gradient method called REINFORCE [11] to update the parameters of the controller [6][12][13]. Policy gradient approaches like Proximal Policy Optimization (PPO) update rule [14] have also been applied to train the sampled architectures and update the controller searching in cell-based search spaces [9], which will be mentioned later in section Ⅲ. Monte Carlo method [15] is another alternative way to search optimal structures. One of the popular algorithms in Monte Carlo methods is the UCT algorithm [16], which has been applied in several NAS research works [17]. TABLE Ⅰ summarizes some of the NAS jobs based on different reinforcement learning algorithms mentioned in this paper.

TABLE I. Neural Architecture Search Based On Different Reinforcement Learning Algorithms

| Frame work | Reinforcement Learning Algorithms Involved |
|---|---|
| NAS[6] | REINFORCE |
| MetaQNN[10] | Q-learning |
| BlockQNN[2] | Q-learning + epsilon-greedy strategy |
| NASNet[3] | Proximal Policy Optimization |
| EAS[12] | Policy Gradient |
| ENAS[18] | REINFORCE |
| TreeCell[13] | REINFORCE |

### A. Neural Architecture Search

NAS uses a recurrent neural network as a controller to sample from a search space to generate new convolutional neural networks. The controller is trained with reinforcement learning. Specifically, the controller acts as an agent to take actions to maximize the rewards by decide which description string is generated in each recurrent network layer. The generated convolutional neural network is built according to the description strings output at each time step by the controller. The rewards are the expected accuracy of the generated convolutional neural networks evaluated in a specific dataset. REINFORCE algorithm was applied as the optimal method to update the parameters of the controller RNN in light of the indifferentiability of the reward. The reinforcement learning algorithm is used to train the controller to make it generate models that can achieve higher accuracy in validation datasets such as CIFAR-10. Later, other researchers used the Proximal Policy Optimization (PPO) which is more sample-efficient than REINFORCE to modify this work to perform faster and more stable in other environments [19]. NAS beat human-designed networks with similar network architectures in CIFAR-10 and achieved new SOTA level in PTB dataset. More than that, NAS found more optimal architecture than widely-used LSTM. The main drawback of the NAS is that its training process needs 800 GPUs which is too costly for other researchers to implement it [6].

*B. MetaQNN*

The other one of the first two pioneering researches that combined neural architecture search with reinforcement learning is MetaQNN. It models neural architecture search problems as Markov Decision process. The main procedures of MetaQNN are very similar with NAS which is mentioned above. What is different is that the reward in MetaQNN is used to train the Q-learning algorithm. The researchers trained NAS in datasets like SVHN, CIFAR-10 and MNIST with 10 GPUs for 8 to 10 days. The best performance of the generated neural network can beat the same-scale networks which are build by human experts [10].

III. IMPROVEMENT IN THE SPEED AND COMPUTATIONAL COST

The two pioneering works of NAS with RL [6][10] need too much computational resource and are trained for too long. On the one hand, the two models need to train the generated models on the validation dataset from scratch every time in order to get accuracy as the reward of the reinforcement learning algorithms. And on the other hand, search space is too large when NAS search the optimal structures for the whole architecture. The speed and computational cost are the main obstructions for the NAS with RL to be experimented and applied widely. To solve these problems, more studies have endeavored to improve the speed and reduce the computational cost have been published. Weight sharing is one of the popular directions of improvement, which is aimed to reuse the weights of existed networks. Hierarchical representations can also accelerate the searching process by reducing the scale of search space.

*A. Weight sharing*

The researchers were inspired by Network Morphism, a method of transform the network without changing its function [20] to reuse the previously trained weights instead of training the weights from scratch. The Efficient Architecture Search (EAS) generates new networks which can represent the same functions as the given networks and are reparameterized to improve the performance. In this case, there is no need for the new networks to be trained from scratch. Hence, it can greatly accelerate the training speed of the generated networks. The meta-controller of EAS functions as a reinforcement learning agent, which takes actions for network transformation [12].

As mentioned above, the Efficient Architecture Search used a function-preserving transformation method called Net2Net [20] which allowed the weight sharing [12]. To address the limitation of classical Network Morphism which can only add or remove layers instead of changing the topology of connected paths, the researchers of TreeCell presented a new type of transformation operations for neural networks called path-level network transformation operations, which allowed modifying the path topology of a certain network while allowing weight sharing to maintain the functionality just as Net2Net operations do[13].

Instead of transforming the smaller networks to larger ones, another method is proposed to select a smaller network from a large and comprehensive one. The process of searching the optimal architecture in classical NAS can be viewed as choosing the optimal path from a single directed acyclic graph (DAG). The Efficient Neural Architecture Search (ENAS) trained selecting a subset of edges within DAG representing a large model simultaneously with the controller. The controller, a long short-term memory(LSTM) network with 100 hidden units was trained to select a path that can maximize the expected reward. One of the advantage of this method is that it enabled the parameters among all architecture in the search space to be shared efficiently. Apart from that, the design of the search space of ENAS enables it to not only learn the operations as NAS [6] did but also design the topology, which is more flexible [18].

*B. Hierarchical Representation*

Due to the fact that the deep neural networks started to include repeated sub-structures called cells or blocks in recent years, researchers have been considering searching based on cells, which means that NAS only search the structures of the cells. But how to connect these cells is pre-defined [1][2][3]. The search space is therefore reduced to include cells of identical structure but different weights.

The researchers of NASNet [3] designed a search space called NASNet search space to make the complexity of the architecture unrelated with the depth of the network and the size of input images. Searching for optimal blocks or cells not only accelerates the speed of training, but also improves the generalization ability of the generated models. Faster BlockQNN was further proposed with network performance prediction [1].

*C. Performance Prediction*

The most time-consuming process of NAS is training the generated networks, which is aimed to evaluate the accuracy of the generated networks. To get the accuracy of the networks more time-efficiently, proxy metrics were used as the approximation of the accuracy. For example, the accuracy of networks training in some smaller datasets or training for less epochs can function as an approximation [3]. Though it may commonly underestimate the true accuracy of models, the target of this process is comparing performance among different networks instead of obtaining the absolute metrics. Due to its efficiency, performance prediction has become more significant in NAS.

From another perspective, it is possible to predict the performance of the networks directly based on the structure of the models. A surrogate model was built to guide the search of network structures [21]. The input of the surrogate model is the string representation of network structures, and the output is the predicted validation accuracy of the generated networks.

Based on an intuitional perspective that many metrics curves are often observed to help researchers decide whether the model is good or not from an overall perspective after training a network for specific epochs. Extrapolation was applied to predict the learning curves [22][23]. After that, Bayesian neural network was used to model and predict learning curves [24].

In TABLE Ⅱ and TABLE Ⅲ, several representative works that have incorporated reinforcement learning with neural architecture search are compared in terms of the performance of the best architectures they generated evaluated on CIFAR-10 and ImageNet. Efficient architecture search costs the least computational resource which is at least 10 times less than any other models evaluated on CIFAR-10. TreeCell obtains the best accuracy with relatively acceptable number of parameters and computational cost. NASNet is the most light-weighted models of all on both datasets and achieves the highest accuracy on ImageNet dataset. BlockQNN can reach satisfactory results with much less time spending on the computation on ImageNet. In general, jobs on improvement of speed and computational cost have fulfilled their aims.

TABLE II. COMPARISON OF SPEED AND RESOURCES ON CIFAR-10

| Models | Performance | | |
|---|---|---|---|
| | Error(%) | # of Params(Millions) | GPU Days |
| MetaQNN[10] | 6.92 | 11.18 | 100 |
| NAS[6] | 3.65 | 37.4 | 22400 |
| EAS[12] | 4.23 | 23.4 | **10** |
| NASNet[3] | 3.41 | **3.3** | 2000 |
| BlockQNN[2] | 3.54 | 39.8 | 96 |
| TreeCell[13] | **2.99** | 5.7 | 200 |

TABLE III. COMPARISON OF SPEED AND RESOURCES ON IMAGENET

| Models | Performance | | | |
|---|---|---|---|---|
| | Top 1 / Top 5 Accuracy (%) | # of Params (Millions) | Image Size (squared) | GPU Days |
| NASNet[3] | **82.7/96.2** | 88.9 | 331 | 2000 |
| BlockQNN[2] | 77.4/93.5 | N/A | 224 | **96** |
| TreeCell[13] | 74.6/91.9 | 594 | 224 | 200 |

IV. EXTENSIONS AND VARIANTS OF NEURAL ARCHITECTURE WITH REINFORCEMENT LEARNING

A. Multi-objective NAS

With the emerge of the need of applying artificial intelligence on platform devices such as mobile phones, more light-weighted networks which are suitable for environments with limited resources such as MobileNet and ShuffleNet started to be proposed and widely researched. NAS also evolved from single objective which only considers accuracy to multiple objectives which consider accuracy, compute intensity, memory, power consumption, latency and so on. However, one of the challenges of the multi-tasks optimization problems is that single solution which can reach the optimal situation of all the subtask simultaneously is hard to be found. In consideration of that, researchers often search for Pareto-optimal solutions.

Mobile neural architecture search (MNAS) approach [25] takes model latency into consideration of the main objective so that the search can identify a model that balances well between accuracy and latency. Resource-efficient neural architect (RENA) [26] takes computational resource use into consideration of automated NAS targets. RENA can find novel architectures that perform competitively even with tight resource constraints. Multi-objective neural architectural search (MONAS) [27] optimizes both accuracy and other objectives that are brought about by numerous devices such as embedded systems, mobile devices and workstations.

V. CONCLUSION

Although the automation of deep learning can generate architectures that perform better than the state-of-the-art hand-crafted networks, the gap is not as large as it is supposed to. One reason for this is too much restrictions that are imposed on the search space. Common search space consists of existing human-designed blocks such as convolutional layers and pooling layers. It is less possible for the neural architecture search to automatically generate novel building blocks. It may increase the performance of the automatically generated models substantially by reducing the limitations on the search space. Hence, the searching for innovative elements of new architectures will be desirable.

REFERENCES

[1] Z. Zhong, Z. Yang, B. Deng, J. Yan, W. Wu, J. Shao and C. L. Liu, 2020. Blockqnn: Efficient block-wise neural network architecture generation. IEEE transactions on pattern analysis and machine intelligence, 43(7), pp.2314-2328.

[2] Z. Zhong, J. Yan, W. Wu, J. Shao and C. L. Liu, 2018. Practical block-wise neural network architecture generation. In *Proceedings of the IEEE conference on computer vision and pattern recognition* (pp. 2423-2432).

[3] B. Zoph, V. Vasudevan, J. Shlens and Q. V. Le, 2018. Learning transferable architectures for scalable image recognition. In *Proceedings of the IEEE conference on computer vision and pattern recognition* (pp. 8697-8710).

[4] L. Li and A. Talwalkar, 2020, August. Random search and reproducibility for neural architecture search. In *Uncertainty in artificial intelligence* (pp. 367-377). PMLR.

[5] T. Elsken, J. H. Metzen and F. Hutter, 2019. Neural architecture search: A survey. *The Journal of Machine Learning Research*, *20*(1), pp.1997-2017.

[6] B. Zoph and Q. V. Le, 2016. Neural architecture search with reinforcement learning. *arXiv preprint arXiv:1611.01578*.

[7] E. Real, S. Moore, A. Selle, S. Saxena, Y. L. Suematsu, J. Tan, Q. V. Le and A. Kurakin, 2017, July. Large-scale evolution of image classifiers. In *International Conference on Machine Learning* (pp. 2902-2911). PMLR.

[8] H. Liu, K. Simonyan and Y. Yang, 2018. Darts: Differentiable architecture search. *arXiv preprint arXiv:1806.09055*.

[9] C. J. C. H. Watkins, 1989. Learning from delayed rewards.

[10] B. Baker, O. Gupta, N. Naik and R. Raskar, 2016. Designing neural network architectures using reinforcement learning. *arXiv preprint arXiv:1611.02167*.

[11] R. J. Williams, 1992. Simple statistical gradient-following algorithms for connectionist reinforcement learning. *Machine learning*, *8*(3), pp.229-256.

[12] H. Cai, T. Chen, W. Zhang, Y. Yu and J. Wang, 2018, April. Efficient architecture search by network transformation. In *Proceedings of the AAAI Conference on Artificial Intelligence* (Vol. 32, No. 1).

[13] H. Cai, J. Yang, W. Zhang, S. Han and Y. Yu, 2018, July. Path-level network transformation for efficient architecture search. In *International Conference on Machine Learning* (pp. 678-687). PMLR.

[14] J. Schulman, F. Wolski, P. Dhariwal, A. Radford and O. Klimov, 2017. Proximal policy optimization algorithms. *arXiv preprint arXiv:1707.06347*.

[15] A. M. Andrew, 1999. REINFORCEMENT LEARNING: AN INTRODUCTION by Richard S. Sutton and Andrew G. Barto, Adaptive Computation and Machine Learning series, MIT Press (Bradford Book), Cambridge, Mass., 1998, xviii+ 322 pp, ISBN 0-262-19398-1,(hardback,£ 31.95). *Robotica*, *17*(2), pp.229-235.

[16] L. Kocsis and C. Szepesvári, 2006, September. Bandit based monte-carlo planning. In *European conference on machine learning* (pp. 282-293). Springer, Berlin, Heidelberg.

[17] R. Negrinho and G. Gordon, 2017. Deeparchitect: Automatically designing and training deep architectures. *arXiv preprint arXiv:1704.08792*.

[18] H. Pham, M. Guan, B. Zoph, Q. Le and J. Dean, 2018, July. Efficient neural architecture search via parameters sharing. In *International conference on machine learning* (pp. 4095-4104). PMLR.

[19] A. Nagy and Á. Boros, 2021. Improving the sample-efficiency of neural architecture search with reinforcement learning. *arXiv preprint arXiv:2110.06751*.

[20] T. Chen, I. Goodfellow and J. Shlens, 2015. Net2net: Accelerating learning via knowledge transfer. *arXiv preprint arXiv:1511.05641*.

[21] C. Liu, B. Zoph, M. Neumann, J. Shlens, W. Hua, L. J. Li, Fei-Fei Li, A. Yuille, J. Huang and K. Murphy, 2018. Progressive neural architecture search. In *Proceedings of the European conference on computer vision (ECCV)* (pp. 19-34).

[22] T. Domhan, J. T. Springenberg and F. Hutter, 2015, June. Speeding up automatic hyperparameter optimization of deep neural networks by extrapolation of learning curves. In *Twenty-fourth international joint conference on artificial intelligence*.

[23] B. Baker, O. Gupta, R. Raskar and N. Naik, 2017. Accelerating neural architecture search using performance prediction. *arXiv preprint arXiv:1705.10823*.

[24] A. Klein, S. Falkner, J. T. Springenberg and F. Hutter, 2016. Learning curve prediction with Bayesian neural networks.

[25] M. Tan, B. Chen, R. Pang, V. Vasudevan, M. Sandler, A. Howard and Q. V. Le, 2019. Mnasnet: Platform-aware neural architecture search for mobile. In *Proceedings of the IEEE/CVF Conference on Computer Vision and Pattern Recognition* (pp. 2820-2828).

[26] Y. Zhou, S. Ebrahimi, S. Ö. Arık, H. Yu, H. Liu and G. Diamos, 2018. Resource-efficient neural architect. *arXiv preprint arXiv:1806.07912*.

[27] C. H. Hsu, S. H. Chang, J. H. Liang, H. P. Chou, C. H. Liu, S. C. Chang, J. Y. Pan, Y. T. Chen, W. Wei and D. C. Juan, 2018. Monas: Multi-objective neural architecture search using reinforcement learning. *arXiv preprint arXiv:1806.10332*.